\documentclass{article} 
\usepackage{iclr2016_conference,times}
\usepackage{hyperref}
\usepackage{url}
\usepackage{graphicx}
\graphicspath{ {./} }

\title{Unsupervised Representation Learning\\ with Deep Convolutional\\ Generative Adversarial Networks}

\author{Alec Radford \& Luke Metz\\
indico Research\\
Boston, MA\\
\texttt{\{alec,luke\}@indico.io}
\AND
Soumith Chintala\\
Facebook AI Research\\
New York, NY\\
\texttt{soumith@fb.com}
}

\begin{document}

\maketitle

\begin{abstract}
In recent years, supervised learning with convolutional networks (CNNs) has seen huge adoption in computer vision applications. Comparatively, unsupervised learning with CNNs has received less attention. In this work we hope to help bridge the gap between the success of CNNs for supervised learning and unsupervised learning. We introduce a class of CNNs called deep convolutional generative adversarial networks (DCGANs), that have certain architectural constraints,  and demonstrate that they are a strong candidate for unsupervised learning. Training on various image datasets, we show convincing evidence that our deep convolutional adversarial pair learns a hierarchy of representations from object parts to scenes in both the generator and discriminator. Additionally, we use the learned features for novel tasks - demonstrating their applicability as general image representations.
\end{abstract}

\section{Introduction}

Learning reusable feature representations from large unlabeled datasets has been an area of active research. In the context of computer vision, one can leverage the practically unlimited amount of unlabeled images and videos to learn good intermediate representations, which can then be used on a variety of supervised learning tasks such as image classification. We propose that one way to build good image representations is by training Generative Adversarial Networks (GANs) \citep{Goodfellow2014}, and later reusing parts of the generator and discriminator networks as feature extractors for supervised tasks. GANs provide an attractive alternative to maximum likelihood techniques. One can additionally argue that their learning process and the lack of a heuristic cost function (such as pixel-wise independent mean-square error) are attractive to representation learning. GANs have been known to be unstable to train, often resulting in generators that produce nonsensical outputs. There has been very limited published research in trying to understand and visualize what GANs learn, and the intermediate representations of multi-layer GANs.

In this paper, we make the following contributions
\begin{itemize}  
    \item We propose and evaluate a set of constraints on the architectural topology of Convolutional GANs that make them stable to train in most settings. We name this class of architectures Deep Convolutional GANs (DCGAN)
    \item We use the trained discriminators for image classification tasks, showing competitive performance with other unsupervised algorithms.
    \item We visualize the filters learnt by GANs and empirically show that specific filters have learned to draw specific objects.
    \item We show that the generators have interesting vector arithmetic properties allowing for easy manipulation of many semantic qualities of generated samples.
\end{itemize}

\section{Related Work}
\subsection{Representation Learning from unlabeled data}
Unsupervised representation learning is a fairly well studied problem in general computer vision research, as well as in the context of images. A classic approach to unsupervised representation learning is to do clustering on the data (for example using K-means), and leverage the clusters for improved classification scores. In the context of images, one can do hierarchical clustering of image patches \citep{coates2012learning} to learn powerful image representations. Another popular method is to train auto-encoders (convolutionally, stacked \citep{vincent2010stacked}, separating the what and where components of the code \citep{zhao2015stacked}, ladder structures \citep{rasmus2015semi}) that encode an image into a compact code, and decode the code to reconstruct the image as accurately as possible. These methods have also been shown to learn good feature representations from image pixels. Deep belief networks \citep{lee2009convolutional} have also been shown to work well in learning hierarchical representations.

\subsection{Generating natural images}

Generative image models are well studied and fall into two categories: parametric and non-parametric.

The non-parametric models often do matching from a database of existing images, often matching patches of images, and have been used in texture synthesis \citep{efros1999texture}, super-resolution \citep{freeman2002example} and in-painting \citep{hays2007scene}.

Parametric models for generating images has been explored extensively (for example on MNIST digits or for texture synthesis \citep{portilla2000parametric}). 
However, generating natural images of the real world have had not much success until recently. A variational sampling approach to generating images \citep{kingma2013auto} has had some success, but the samples often suffer from being blurry. Another approach generates images using an iterative forward diffusion process \citep{sohl2015deep}. Generative Adversarial Networks \citep{Goodfellow2014} generated images suffering from being noisy and incomprehensible. A laplacian pyramid extension to this approach \citep{denton2015deep} showed higher quality images, but they still suffered from the objects looking wobbly because of noise introduced in chaining multiple models. A recurrent network approach \citep{gregor2015draw} and a deconvolution network approach \citep{dosovitskiy2014learning} have also recently had some success with generating natural images. However, they have not leveraged the generators for supervised tasks.

\subsection{Visualizing the internals of CNNs}

One constant criticism of using neural networks has been that they are black-box methods, with little understanding of what the networks do in the form of a simple human-consumable algorithm. In the context of CNNs, Zeiler et. al. \citep{zeiler2014visualizing} showed that by using deconvolutions and filtering the maximal activations, one can find the approximate purpose of each convolution filter in the network. Similarly, using a gradient descent on the inputs lets us inspect the ideal image that activates certain subsets of filters \citep{Inceptionism2015}. 

\section{Approach and Model Architecture}

Historical attempts to scale up GANs using CNNs to model images have been unsuccessful. This motivated the authors of LAPGAN \citep{denton2015deep} to develop an alternative approach to iteratively upscale low resolution generated images which can be modeled more reliably. We also encountered difficulties attempting to scale GANs using CNN architectures commonly used in the supervised literature. However, after extensive model exploration we identified a family of architectures that resulted in stable training across a range of datasets and allowed for training higher resolution and deeper generative models.

Core to our approach is adopting and modifying three recently demonstrated changes to CNN architectures. 

The first is the all convolutional net \citep{springenberg2014striving} which replaces deterministic spatial pooling functions (such as maxpooling) with strided convolutions, allowing the network to learn its own spatial downsampling. We use this approach in our generator, allowing it to learn its own spatial upsampling, and discriminator.

\begin{figure}[h]
\begin{center}
\includegraphics[width=13.9cm]{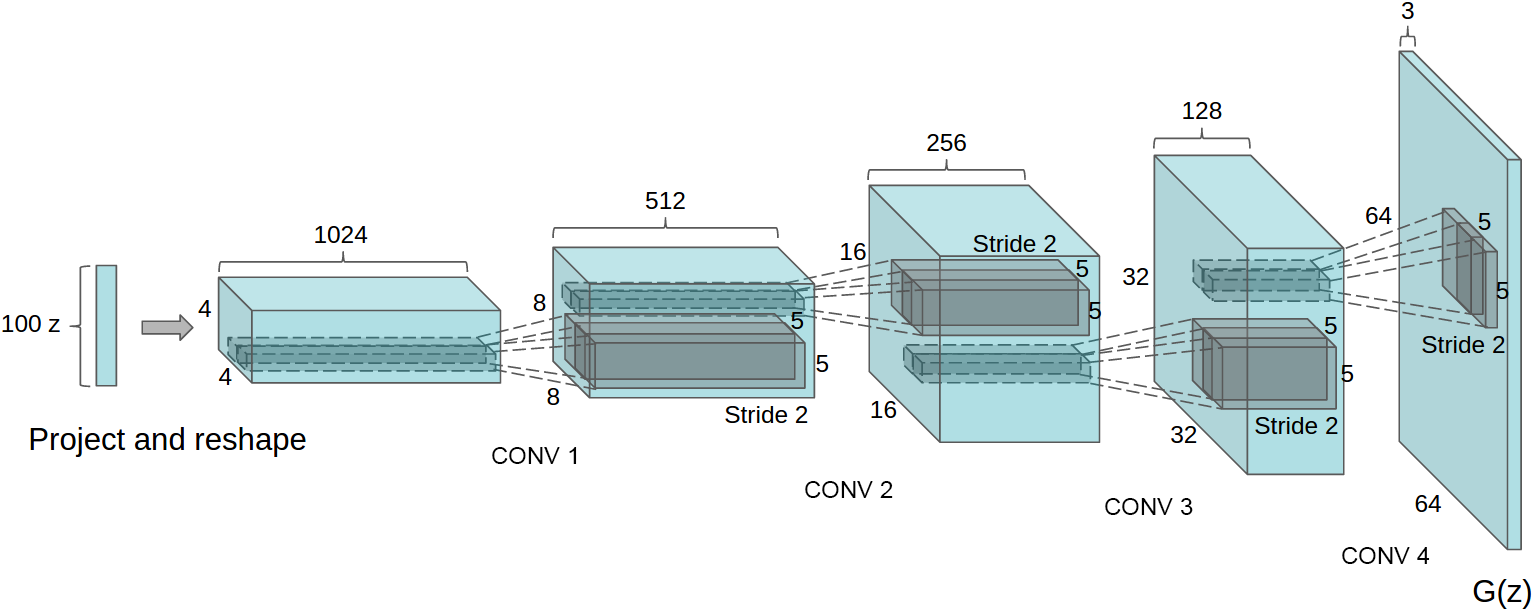}
\end{center}
\caption{\label{fig_LSUN_architecture}DCGAN generator used for LSUN scene modeling. A 100 dimensional uniform distribution $Z$ is projected to a small spatial extent convolutional representation with many feature maps. A series of four fractionally-strided convolutions (in some recent papers, these are wrongly called deconvolutions) then convert this high level representation into a $64 \times 64$ pixel image. Notably, no fully connected or pooling layers are used.}
\end{figure}

Second is the trend towards eliminating fully connected layers on top of convolutional features. The strongest example of this is global average pooling which has been utilized in state of the art image classification models \citep{Inceptionism2015}. We found global average pooling increased model stability but hurt convergence speed. A middle ground of directly connecting the highest convolutional features to the input and output respectively of the generator and discriminator worked well. The first layer of the GAN, which takes a uniform noise distribution $Z$ as input, could be called fully connected as it is just a matrix multiplication, but the result is reshaped into a 4-dimensional tensor and used as the start of the convolution stack. For the discriminator, the last convolution layer is flattened and then fed into a single sigmoid output. See Fig. \ref{fig_LSUN_architecture} for a visualization of an example model architecture.

Third is Batch Normalization \citep{ioffe2015batch} which stabilizes learning by normalizing the input to each unit to have zero mean and unit variance. This helps deal with training problems that arise due to poor initialization and helps gradient flow in deeper models. This proved critical to get deep generators to begin learning, preventing the generator from collapsing all samples to a single point which is a common failure mode observed in GANs. Directly applying batchnorm to all layers however, resulted in sample oscillation and model instability. This was avoided by not applying batchnorm to the generator output layer and the discriminator input layer.

The ReLU activation \citep{nair2010rectified} is used in the generator with the exception of the output layer which uses the Tanh function. We observed that using a bounded activation allowed the model to learn more quickly to saturate and cover the color space of the training distribution. Within the discriminator we found the leaky rectified activation \citep{maas2013rectifier} \citep{xu2015empirical} to work well, especially for higher resolution modeling. This is in contrast to the original GAN paper, which used the maxout activation \citep{goodfellow2013maxout}.

\fbox{
    \parbox{\textwidth}
    {
        Architecture guidelines for stable Deep Convolutional GANs
        \begin{itemize}
            \item Replace any pooling layers with strided convolutions (discriminator) and fractional-strided convolutions (generator).
            \item Use batchnorm in both the generator and the discriminator.
            \item Remove fully connected hidden layers for deeper architectures.
            \item Use ReLU activation in generator for all layers except for the output, which uses Tanh.
            \item Use LeakyReLU activation in the discriminator for all layers.
        \end{itemize}
    }
}

\section{Details of adversarial training}

We trained DCGANs on three datasets, Large-scale Scene Understanding (LSUN) \citep{yu2015construction}, Imagenet-1k and a newly assembled Faces dataset. Details on the usage of each of these datasets are given below.

No pre-processing was applied to training images besides scaling to the range of the tanh activation function [-1, 1]. All models were trained with mini-batch stochastic gradient descent (SGD) with a mini-batch size of 128. All weights were initialized from a zero-centered Normal distribution with standard deviation 0.02. In the LeakyReLU, the slope of the leak was set to 0.2 in all models. While previous GAN work has used momentum to accelerate training, we used the Adam optimizer \citep{kingma2014adam} with tuned hyperparameters. We found the suggested learning rate of 0.001, to be too high, using 0.0002 instead. Additionally, we found leaving the momentum term $\beta_1$ at the suggested value of 0.9 resulted in training oscillation and instability while reducing it to 0.5 helped stabilize training.

\subsection{LSUN}

\begin{figure}[h]
\begin{center}
\includegraphics[width=13.9cm]{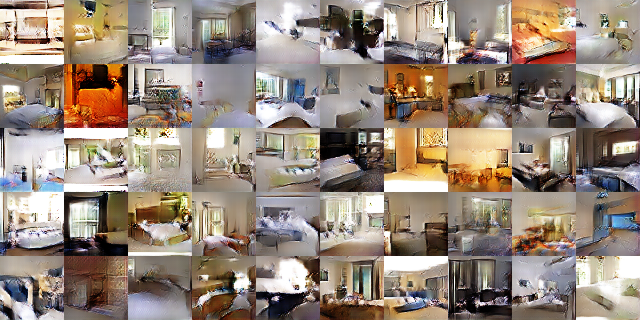}
\end{center}
\caption{\label{fig_LSUN_one_epoch} Generated bedrooms after one training pass through the dataset. Theoretically, the model could learn to memorize training examples, but this is experimentally unlikely as we train with a small learning rate and minibatch SGD. We are aware of no prior empirical evidence demonstrating memorization with SGD and a small learning rate.}
\end{figure}

\begin{figure}[h]
\begin{center}
\includegraphics[width=13.9cm]{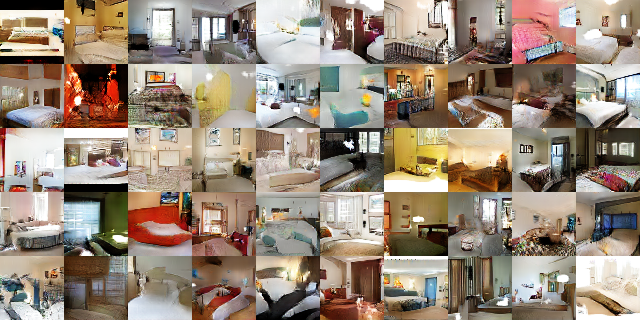}
\end{center}
\caption{\label{fig_LSUN_five_epoch} Generated bedrooms after five epochs of training. There appears to be evidence of visual under-fitting via repeated noise textures across multiple samples such as the base boards of some of the beds.}
\end{figure}

As visual quality of samples from generative image models has improved, concerns of over-fitting and memorization of training samples have risen. 
To demonstrate how our model scales with more data and higher resolution generation, we train a model on the LSUN bedrooms dataset  containing a little over 3 million training examples. Recent analysis has shown that there is a direct link between how fast models learn and their generalization performance \citep{hardt2015train}. We show samples from one epoch of training (Fig.\ref{fig_LSUN_one_epoch}), mimicking online learning, in addition to samples after convergence (Fig.\ref{fig_LSUN_five_epoch}), as an opportunity to demonstrate that our model is not producing high quality samples via simply overfitting/memorizing training examples. No data augmentation was applied to the images.

\subsubsection{Deduplication}
To further decrease the likelihood of the generator memorizing input examples (Fig.\ref{fig_LSUN_one_epoch}) we perform a simple image de-duplication process. We fit a 3072-128-3072 de-noising dropout regularized RELU autoencoder on 32x32 downsampled center-crops of training examples. The resulting code layer activations are then binarized via thresholding the ReLU activation which has been shown to be an effective information preserving technique \citep{srivastava2014understanding} and provides a convenient form of semantic-hashing, allowing for linear time de-duplication . Visual inspection of hash collisions showed high precision with an estimated false positive rate of less than 1 in 100. Additionally, the technique detected and removed approximately 275,000 near duplicates, suggesting a high recall.

\subsection{Faces}
We scraped images containing human faces from random web image queries of peoples names. The people names were acquired from dbpedia, with a criterion that they were born in the modern era. This dataset has ~3M images from 10K people. We run an OpenCV face detector on these images, keeping the detections that are sufficiently high resolution, which gives us approximately 350,000 face boxes. We use these face boxes for training. No data augmentation was applied to the images.

\subsection{Imagenet-1k}
We use Imagenet-1k \citep{deng2009imagenet} as a source of natural images for unsupervised training. We train on $32 \times 32$ min-resized center crops. No data augmentation was applied to the images.

\section{Empirical Validation of DCGANs capabilities}

\subsection{Classifying CIFAR-10 using GANs as a feature extractor}

One common technique for evaluating the quality of unsupervised representation learning algorithms is to apply them as a feature extractor on supervised datasets and evaluate the performance of linear models fitted on top of these features. 

On the CIFAR-10 dataset, a very strong baseline performance has been demonstrated from a well tuned single layer feature extraction pipeline utilizing K-means as a feature learning algorithm. When using a very large amount of feature maps (4800) this technique achieves 80.6\% accuracy. An unsupervised multi-layered extension of the base algorithm reaches 82.0\% accuracy \citep{coates2011selecting}. To evaluate the quality of the representations learned by DCGANs for supervised tasks, we train on Imagenet-1k and then use the discriminator's convolutional features from all layers, maxpooling each layers representation to produce a $4 \times 4$ spatial grid. These features are then flattened and concatenated to form a 28672 dimensional vector and a regularized linear L2-SVM classifier is trained on top of them. This achieves 82.8\% accuracy, out performing all K-means based approaches. Notably, the discriminator has many less feature maps (512 in the highest layer) compared to K-means based techniques, but does result in a larger total feature vector size due to the many layers of $4 \times 4$ spatial locations. The performance of DCGANs is still less than that of Exemplar CNNs \citep{dosovitskiy2014discriminative}, a technique which trains normal discriminative CNNs in an unsupervised fashion to differentiate between specifically chosen, aggressively augmented, exemplar samples from the source dataset. Further improvements could be made by finetuning the discriminator's representations, but we leave this for future work. Additionally, since our DCGAN was never trained on CIFAR-10 this experiment also demonstrates the domain robustness of the learned features. 

\begin{table}[h]
\caption{CIFAR-10 classification results using our pre-trained model. Our DCGAN is not pre-trained on CIFAR-10, but on Imagenet-1k, and the features are used to classify CIFAR-10 images.}
\begin{center}
\begin{tabular}{ |c|c|c|c| }
    \hline
    \textbf{Model} & Accuracy & Accuracy (400 per class) & max \# of features units\\
    \hline
    1 Layer K-means & 80.6\% & 63.7\% ($\pm$0.7\%) & 4800\\
    3 Layer K-means Learned RF & 82.0\% & 70.7\% ($\pm$0.7\%) & 3200\\
    View Invariant K-means & 81.9\% & 72.6\% ($\pm$0.7\%) & 6400\\
    Exemplar CNN & 84.3\% & 77.4\% ($\pm$0.2\%) & 1024\\
    \hline
    DCGAN (ours) + L2-SVM & 82.8\% & 73.8\% ($\pm$0.4\%) & 512\\
    \hline
\end{tabular}
\end{center}
\label{tab:cifar10_results}
\end{table}

\subsection{Classifying SVHN digits using GANs as a feature extractor}

On the StreetView House Numbers dataset (SVHN)\citep{netzer2011reading}, we use the features of the discriminator of a DCGAN for supervised purposes when labeled data is scarce. Following similar dataset preparation rules as in the CIFAR-10 experiments, we split off a validation set of 10,000 examples from the non-extra set and use it for all hyperparameter and model selection. 1000 uniformly class distributed training examples are randomly selected and used to train a regularized linear L2-SVM classifier on top of the same feature extraction pipeline used for CIFAR-10. This achieves state of the art (for classification using 1000 labels) at 22.48\% test error, improving upon another modifcation of CNNs designed to leverage unlabled data \citep{zhao2015stacked}. Additionally, we validate that the CNN architecture used in DCGAN is not the key contributing factor of the model's performance by training a purely supervised CNN with the same architecture on the same data and optimizing this model via random search over 64 hyperparameter trials \citep{bergstra2012hpopt}. It achieves a signficantly higher 28.87\% validation error.

\begin{table}[h]
\caption{SVHN classification with 1000 labels}
\begin{center}
\begin{tabular}{ |c|c| }
    \hline
    \textbf{Model} & error rate\\
    \hline
    KNN & 77.93\%\\
    TSVM & 66.55\%\\
    M1+KNN & 65.63\%\\
    M1+TSVM & 54.33\%\\
    M1+M2 & 36.02\%\\
    \hline
    SWWAE without dropout & 27.83\%\\
    SWWAE with dropout & 23.56\%\\
    \hline
    DCGAN (ours) + L2-SVM & 22.48\%\\
    Supervised CNN with the same architecture & 28.87\% (validation)\\
    \hline
\end{tabular}
\end{center}
\label{tab:svhn_results}
\end{table}

\section{Investigating and visualizing the internals of the networks}

We investigate the trained generators and discriminators in a variety of ways. We do not do any kind of nearest neighbor search on the training set. Nearest neighbors in pixel or feature space are trivially fooled \citep{Theis2015d} by small image transforms. We also do not use log-likelihood metrics to quantitatively assess the model, as it is a poor \citep{Theis2015d} metric. 

\subsection{Walking in the latent space}
\begin{figure}[h!]
\begin{center}
\includegraphics[width=13.9cm]{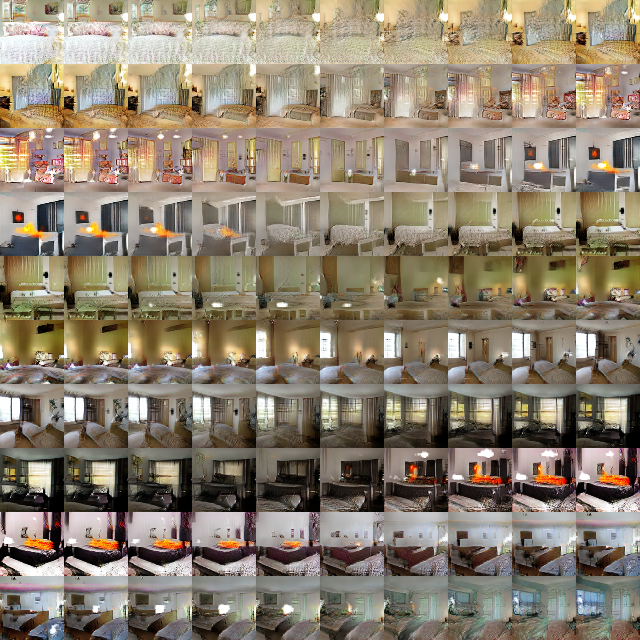}
\end{center}
\caption{\label{fig_LSUN_interp}Top rows: Interpolation between a series of 9 random points in $Z$ show that the space learned has smooth transitions, with every image in the space plausibly looking like a bedroom. In the 6th row, you see a room without a window slowly transforming into a room with a giant window. In the 10th row, you see what appears to be a TV slowly being transformed into a window. }
\end{figure}

The first experiment we did was to understand the landscape of the latent space. Walking on the manifold that is learnt can usually tell us about signs of memorization (if there are sharp transitions) and about the way in which the space is hierarchically collapsed. If walking in this latent space results in semantic changes to the image generations (such as objects being added and removed), we can reason that the model has learned relevant and interesting representations. The results are shown in Fig.\ref{fig_LSUN_interp}.

\subsection{Visualizing the Discriminator Features}

\begin{figure}[h!]
\begin{center}
\includegraphics[width=13.9cm]{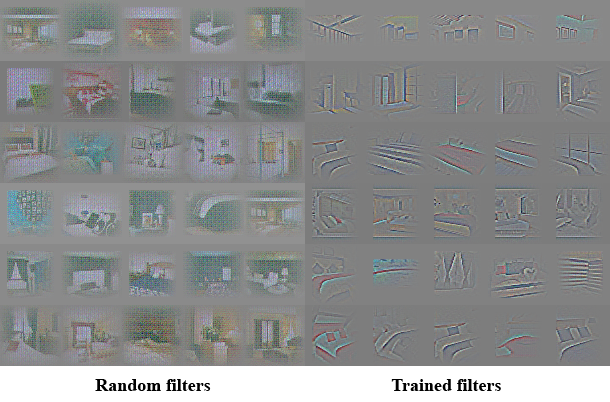}
\end{center}

\caption{\label{fig_LSUN_guidedbp_trained}On the right, guided backpropagation visualizations of maximal axis-aligned responses for the first 6 learned convolutional features from the last convolution layer in the discriminator. Notice a significant minority of features respond to beds - the central object in the LSUN bedrooms dataset. On the left is a random filter baseline. Comparing to the previous responses there is little to no discrimination and random structure.}
\end{figure}

Previous work has demonstrated that supervised training of CNNs on large image datasets results in very powerful learned features \citep{zeiler2014visualizing}.  Additionally, supervised CNNs trained on scene classification learn object detectors \citep{Oquab14}. We demonstrate that an unsupervised DCGAN trained on a large image dataset can also learn a hierarchy of features that are interesting.
Using guided backpropagation as proposed by \citep{springenberg2014striving}, we show in Fig.\ref{fig_LSUN_guidedbp_trained} that the features learnt by the discriminator activate on typical parts of a bedroom, like beds and windows. For comparison, in the same figure, we give a baseline for randomly initialized features that are not activated on anything that is semantically relevant or interesting.

\subsection{Manipulating the Generator Representation}

\subsubsection{Forgetting to draw certain objects}

In addition to the representations learnt by a discriminator, there is the question of what representations the generator learns. The quality of samples suggest that the generator learns specific object representations for major scene components such as beds, windows, lamps, doors, and miscellaneous furniture. In order to explore the form that these representations take, we conducted an experiment to attempt to remove windows from the generator completely. 

\begin{figure}[h!]
\begin{center}
\includegraphics[width=13.9cm]{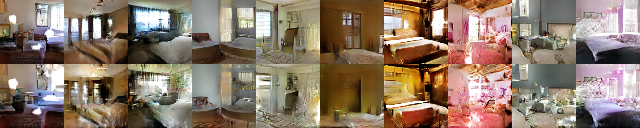}
\end{center}
\caption{\label{fig_LSUN_window_remove}Top row: un-modified samples from model. Bottom row: the same samples generated with dropping out "window" filters. Some windows are removed, others are transformed into objects with similar visual appearance such as doors and mirrors. Although visual quality decreased, overall scene composition stayed similar, suggesting the generator has done a good job disentangling scene representation from object representation. Extended experiments could be done to remove other objects from the image and modify the objects the generator draws.}
\end{figure}

On 150 samples, 52 window bounding boxes were drawn manually. On the second highest convolution layer features, logistic regression was fit to predict whether a feature activation was on a window (or not), by using the criterion that activations inside the drawn bounding boxes are positives and random samples from the same images are negatives. Using this simple model, all feature maps with weights greater than zero (~200 in total) were dropped from all spatial locations. Then, random new samples were generated with and without the feature map removal. 

The generated images with and without the window dropout are shown in Fig.\ref{fig_LSUN_window_remove}, and interestingly, the network mostly forgets to draw windows in the bedrooms, replacing them with other objects.

\begin{figure}[h!]
\begin{center}
\includegraphics[width=13.9cm]{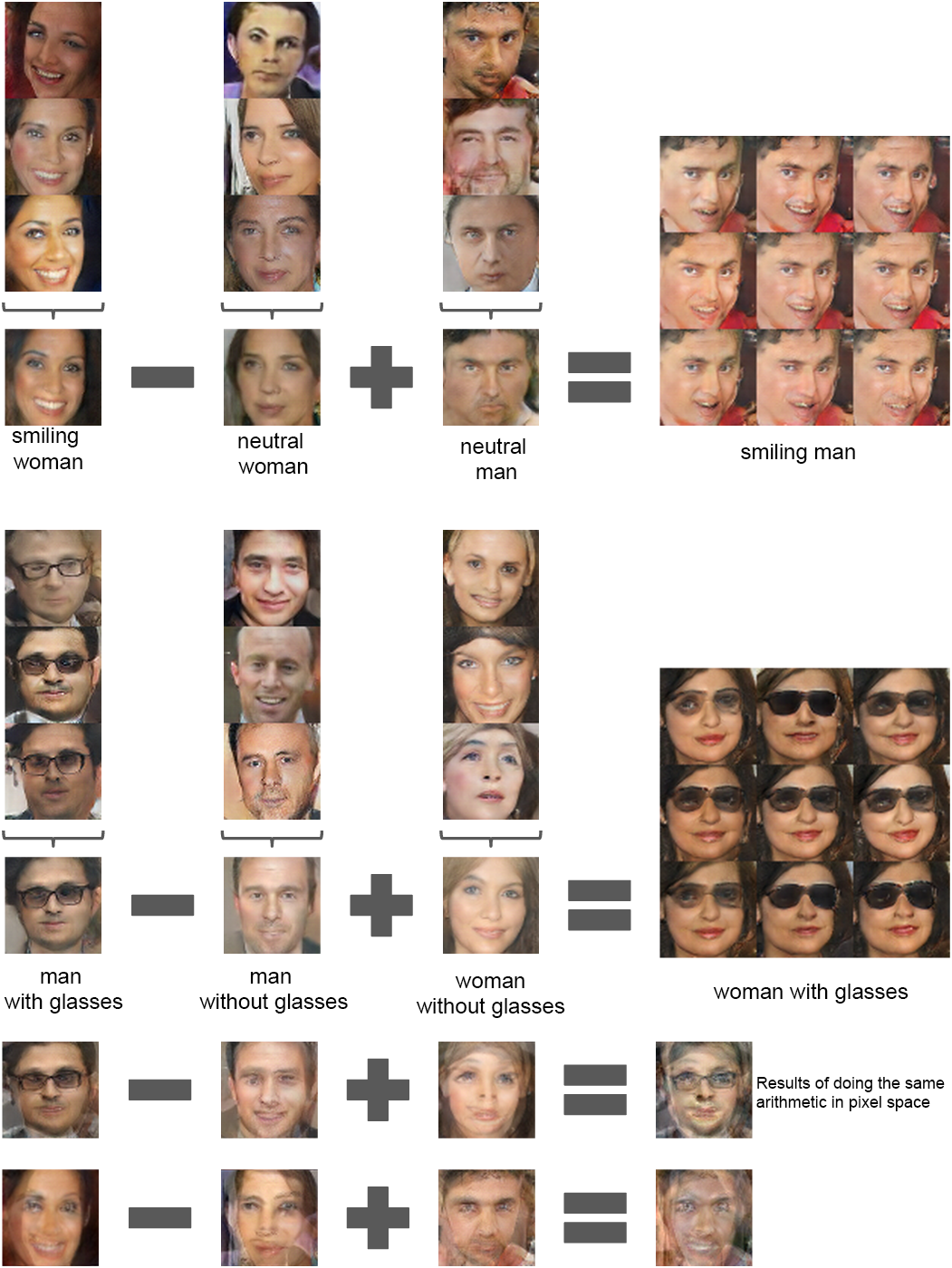}
\end{center}
\caption{\label{fig_faces_vector_arithmetic}Vector arithmetic for visual concepts. For each column, the $Z$ vectors of samples are averaged. Arithmetic was then performed on the mean vectors creating a new vector $Y$. The center sample on the right hand side is produce by feeding $Y$ as input to the generator. To demonstrate the interpolation capabilities of the generator, uniform noise sampled with scale +-0.25 was added to $Y$ to produce the 8 other samples. Applying arithmetic in the input space (bottom two examples) results in noisy overlap due to misalignment.}
\end{figure}

\begin{figure}[h!]
\begin{center}
\includegraphics[width=13.9cm]{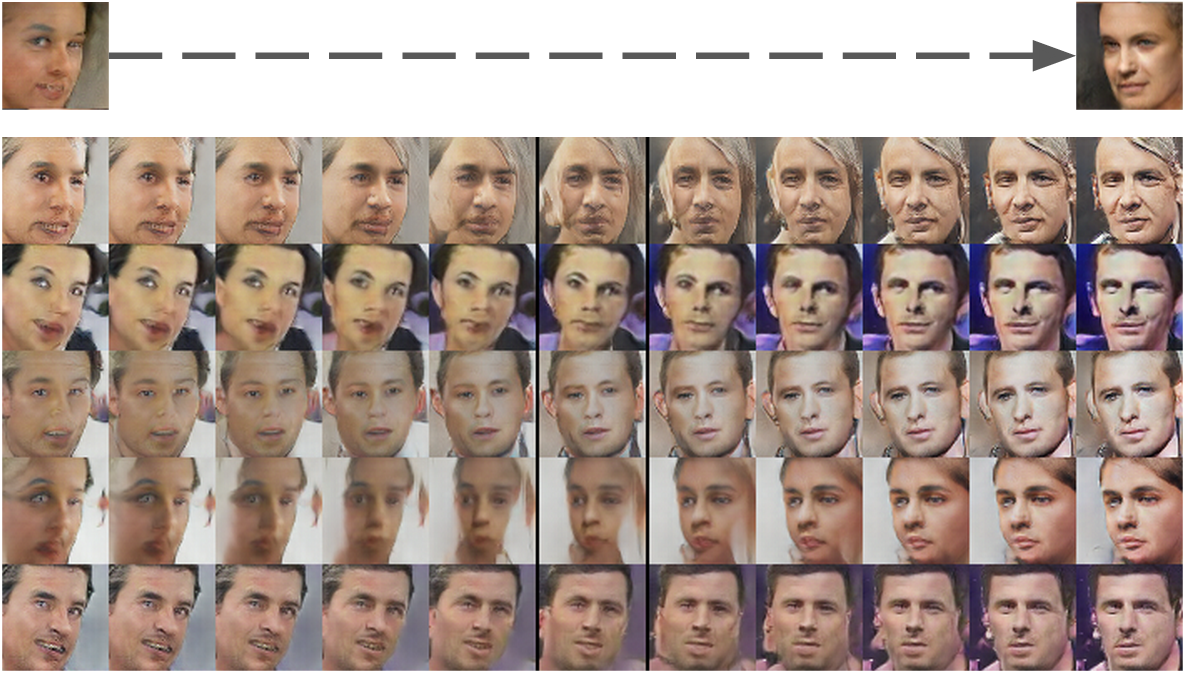}
\end{center}
\caption{\label{fig_faces_vector_turn}A "turn" vector was created from four averaged samples of faces looking left vs looking right. By adding interpolations along this axis to random samples we were able to reliably transform their pose.}
\end{figure}

\subsubsection{Vector arithmetic on face samples}

In the context of evaluating learned representations of words \citep{mikolov2013distributed} demonstrated that simple arithmetic operations revealed rich linear structure in representation space. One canonical example demonstrated that the vector("King") - vector("Man") + vector("Woman") resulted in a vector whose nearest neighbor was the vector for Queen. We investigated whether similar structure emerges in the $Z$ representation of our generators. We performed similar arithmetic on the $Z$ vectors of sets of exemplar samples for visual concepts. Experiments working on only single samples per concept were unstable, but averaging the $Z$ vector for three examplars showed consistent and stable generations that semantically obeyed the arithmetic. In addition to the object manipulation shown in (Fig. \ref{fig_faces_vector_arithmetic}), we demonstrate that face pose is also modeled linearly in $Z$ space (Fig. \ref{fig_faces_vector_turn}).  

These demonstrations suggest interesting applications can be developed using $Z$ representations learned by our models. It has been previously demonstrated that conditional generative models can learn to convincingly model object attributes like scale, rotation, and position \citep{dosovitskiy2014learning}. This is to our knowledge the first demonstration of this occurring in purely unsupervised models. Further exploring and developing the above mentioned vector arithmetic could dramatically reduce the amount of data needed for conditional generative modeling of complex image distributions.

\section{Conclusion and Future Work}

We propose a more stable set of architectures for training generative adversarial networks and we give evidence that adversarial networks learn good representations of images for supervised learning and generative modeling. There are still some forms of model instability remaining - we noticed as models are trained longer they sometimes collapse a subset of filters to a single oscillating mode. Further work is needed to tackle this from of instability. We think that extending this framework to other domains such as video (for frame prediction) and audio (pre-trained features for speech synthesis) should be very interesting. Further investigations into the properties of the learnt latent space would be interesting as well.
\subsubsection*{Acknowledgments}

We are fortunate and thankful for all the advice and guidance we have received during this work, especially that of Ian Goodfellow, Tobias Springenberg, Arthur Szlam and Durk Kingma. Additionally we'd like to thank all of the folks at indico for providing support, resources, and conversations, especially the two other members of the indico research team, Dan Kuster and Nathan Lintz. Finally, we'd like to thank Nvidia for donating a Titan-X GPU used in this work.

\bibliography{iclr2016_conference}
\bibliographystyle{iclr2016_conference}

\newpage

\section{Supplementary Material}

\subsection{Evaluating DCGANs capability to capture data distributions}

We propose to apply standard classification metrics to a conditional version of our model, evaluating the conditional distributions learned. We trained a DCGAN on MNIST (splitting off a 10K validation set) as well as a permutation invariant GAN baseline and evaluated the models using a nearest neighbor classifier comparing real data to a set of generated conditional samples. We found that removing the scale and bias parameters from batchnorm produced better results for both models. We speculate that the noise introduced by batchnorm helps the generative models to better explore and generate from the underlying data distribution. The results are shown in Table ~\ref{tab:nnc_mnist_results} which compares our models with other techniques. The DCGAN model achieves the same test error as a nearest neighbor classifier fitted on the training dataset - suggesting the DCGAN model has done a superb job at modeling the conditional distributions of this dataset. At one million samples per class, the DCGAN model outperforms InfiMNIST \citep{loosli-canu-bottou-2006}, a hand developed data augmentation pipeline which uses translations and elastic deformations of training examples. The DCGAN is competitive with a probabilistic generative data augmentation technique utilizing learned per class transformations \citep{Hauberg2015} while being more general as it directly models the data instead of transformations of the data.

\begin{table}[h]
\caption{Nearest neighbor classification results.}
\begin{center}
\begin{tabular}{ |c|c|c| }
    \hline
    \textbf{Model} & Test Error @50K samples & Test Error @10M samples\\
    \hline
    AlignMNIST & - & 1.4\% \\
    InfiMNIST & - & 2.6\% \\
    Real Data & 3.1\% & - \\
    GAN & 6.28\% & 5.65\% \\
    \hline
    DCGAN (ours) & 2.98\% & 1.48\% \\
    \hline
\end{tabular}
\end{center}
\label{tab:nnc_mnist_results}
\end{table}

\begin{figure}[h]
\begin{center}
\includegraphics[width=13.9cm]{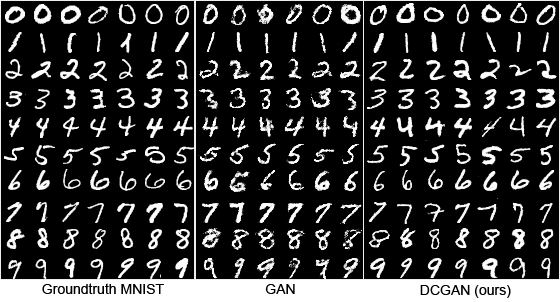}
\end{center}
\caption{\label{fig_mnist_collage} Side-by-side illustration of (from left-to-right) the MNIST dataset, generations from a baseline GAN, and generations from our DCGAN .}
\end{figure}

\begin{figure}[h]
\begin{center}
\includegraphics[width=13.9cm]{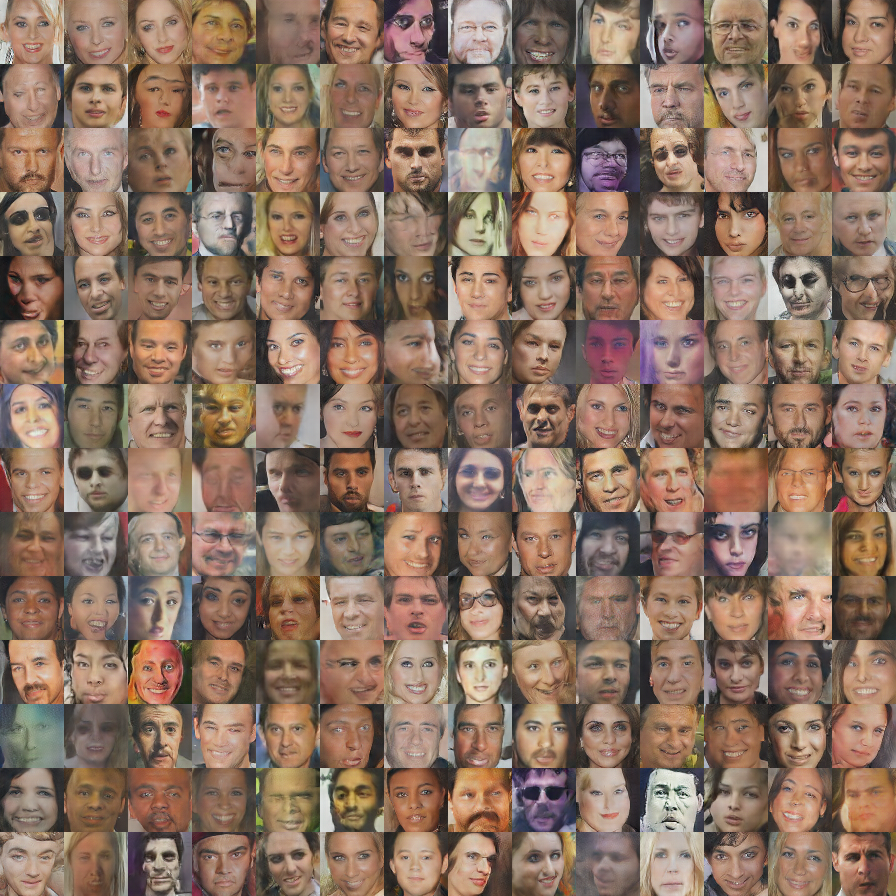}
\end{center}
\caption{\label{fig_albums} More face generations from our Face DCGAN.}
\end{figure}

\begin{figure}[h]
\begin{center}
\includegraphics[width=13.9cm]{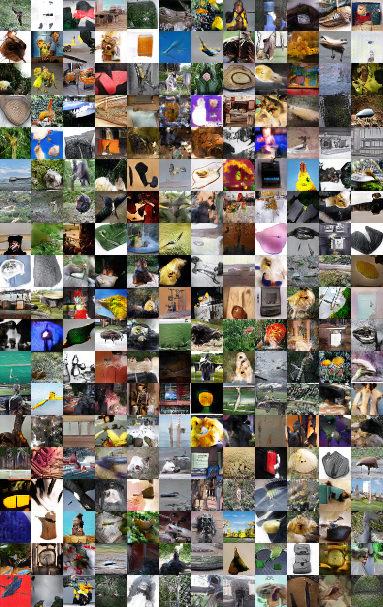}
\end{center}
\caption{\label{fig_imagenet} Generations of a DCGAN that was trained on the Imagenet-1k dataset.}
\end{figure}

\end{document}